\pdfoutput=1

\documentclass[11pt]{article}

\usepackage{acl}
\usepackage{amsmath}
\usepackage{bm}
\usepackage{amsfonts}
\usepackage{booktabs}
\usepackage{multirow}
\usepackage{times}
\usepackage{latexsym}
\usepackage{cleveref}
\usepackage{subfigure}
\newtheorem{theorem}{Theorem}

\newtheorem{definition}{Definition}
\newtheorem{proof}{Proof}

\usepackage[T1]{fontenc}

\usepackage[utf8]{inputenc}

\usepackage{microtype}

\title{Frequency-aware Dimension Selection for Static Word Embedding by Mixed Product Distance}
\author{Lingfeng Shen \\ Johns Hopkins Iniversity
        \And  Haiyun Jiang \\ Tencent AI Lab \And Lemao Liu \\ Tencent AI Lab \And
        Ying Chen \\ China Agricultural University}

\begin{document}
\maketitle
\begin{abstract}
Static word embedding is still useful, particularly for context-unavailable tasks, because in the case of no context available, pre-trained language models often perform worse than static word embeddings.
Although dimension is a key factor determining the quality of static word embeddings, automatic dimension selection is rarely discussed. In this paper, we investigate the impact of word frequency on the dimension selection, and empirically find that word frequency is so vital that it needs to be taken into account during dimension selection. 
Based on such an empirical finding, this paper proposes a dimension selection method that uses a metric (Mixed Product Distance, MPD) to select a proper dimension for word embedding algorithms without training any word embedding. Through applying a post-processing function to oracle matrices, the MPD-based method can de-emphasize the impact of word frequency. Experiments on both context-unavailable and context-available tasks demonstrate the better efficiency-performance trade-off of our MPD-based dimension selection method over baselines.
\end{abstract}
\section{Introduction}
Word embedding has been widely used in numerous NLP tasks, such as recommendation systems \cite{zhang2019deep}, text classification \cite{chung2018speech2vec}, information retrieval \cite{palangi2016deep} and machine translation \cite{guo2019non}. Though pre-trained language models (PLMs) like BERT \cite{devlin2018bert} have become a prevailed paradigm in NLP, static word embeddings, such as Word2Vec \cite{mikolov2013distributed} and GloVe \cite{pennington2014GloVe}, are still useful in scenarios where contexts are missing, such as entity retrieval \cite{liu2021texsmart,han2020case} and word similarity tasks \cite{halawi2012large,bruni2014multimodal}. 
Under such scenarios, static word embeddings often perform better than PLMs. 

Although there are many algorithms for training static word embeddings, the impact of dimension on word embedding has been less investigated.  \citet{yin2018dimensionality} have shown that a critical hyper-parameter, the choice of dimension for word vectors, significantly influences the performance of a model that builds on word embeddings. An extremely low-dimensional word embedding is probably not expressive enough to represent the meanings of words, and in contrast, a high-dimensional word embedding is expressive but possibly leads to over-fitting and high computational cost. For most NLP models building on word embeddings, dimension is selected ad hoc or by grid search, either resulting in sub-optimal model performances. Thus, it is promising to establish a dimension selection criterion that can automatically select a proper dimension for static word embedding algorithms with a good efficiency-performance trade-off.

\citet{yin2018dimensionality} first investigated dimension selection for static word embedding. They proposed a metric called Pairwise Inner Product (PIP) loss which attempts to reflect the relationship between the dimension and the quality of word embedding and then developed a PIP-based dimension selection method that selects a dimension with the minimal PIP loss. 
Later, \citet{wang2019single} proposed a dimension selection method based on Principal Components Analysis (PCA). The PCA-based dimension selection trains a high-dimensional word embedding and then uses the embedding to conduct a grid search to find a proper dimension, which is time-consuming. However, both the PIP-based dimension selection and the PCA-based dimension selection overlook one dominant element that may influence the dimension selection: word frequency.

Thus, in this paper, we first investigate whether word frequency is influential in the dimension selection of static word embedding. 
Our results reveal that extreme word frequency may cause an improper dimension for the PIP-based or PCA-based dimension selection. 
That is, without diminishing the impact of word frequency, it is more likely to select an improper dimension and generate a low-quality word embedding. 
Then, we explore a way to diminish the impact of word frequency. 
Inspired by the success of post-processing methods on static word embeddings \cite{mu2018all,liu2019unsupervised}, which de-emphasize the influence of some words in an embedding, we attempt to transfer them to the dimension selection.

Specifically, we propose an MPD-based dimension selection, which minimizes a metric called \textbf{Mixed Product Distance (MPD)} to find a proper dimension. The MPD can reduce the impact of word frequency by applying a post-processing function to oracle matrices, which results in a more proper dimension. To investigate the effectiveness of our method, we conduct evaluations on context-unavailable and context-available tasks. Comprehensive experiments demonstrate that the MPD-base dimension selection method can consistently perform better than baselines.

Above all, the main contributions of the paper can be outlined as follows:
\begin{itemize} 
    \item We expose a critical issue overlooked by previous dimension selection methods: word frequency. It can influence the dimension selection and result in degraded word embeddings.
    \item We design a metric called Mixed Product Distance (MPD) for static word embedding dimension selection, which uses post-process functions to overcome the word frequency problem.
    \item We evaluate the MPD-based dimension selection on various NLP tasks, including context-unavailable and context-available tasks. Comprehensive experiments demonstrate that our method achieves a better efficiency-performance trade-off. 
\end{itemize}

\section{Related Work}
\subsection{Static Word Embedding and Dimension Selection}
Existing static word embedding algorithms can be formulated as either explicit or implicit matrix factorization, and the most popular word embedding algorithms can be regarded as \textbf{implicit} matrix factorization. For example, Word2Vec and GloVe perform implicit factorization on the symmetric matrix (i.e., $\alpha=0.5$), where the signal matrix $M$ is the shifted PPMI matrix for Word2Vec and the log-count matrix for GloVe. 
Although intensive studies have been done on static word embedding algorithms that learn word embeddings using matrix factorization, selecting a proper dimension for these word embedding algorithms is less investigated. \citet{yin2018dimensionality} proposed a PIP-based dimension selection method. Specifically, for every possible dimension, its PIP loss is obtained, and then the dimension with the minimal PIP loss is selected. Particularly, the PIP-based method can be conducted without training word embedding. In contrast, \citet{wang2019single} proposed a PCA-based method for dimension selection, which trains a high-dimensional word embedding to help the dimension selection, which results in extremely time-consuming.

\subsection{Post-processing for Word Embedding}
To further improve the quality of a word embedding, post-processing which reinforces linguistic constraints on the embedding, is often used. According to the studies on post-processing \cite{mu2018all,liu2019unsupervised}, one of the important linguistic constraints is to diminish the influence of word frequency. Specifically, in a word embedding, principal components (PCs), which are mainly encoded by frequent words, are shared by all words. As a result, the vectors of less frequent words have a higher variance than those of frequent words, and such a high variance hurts the embedding quality because word frequency is unrelated to lexical semantics. Moreover, two popular post-processing functions, the All-but-the-top (ABTT) \cite{mu2018all} and Conceptor Negation (CN) \cite{liu2019unsupervised}, have been used to remove or suppress PCs in word embeddings. In our method, instead of word embedding, these post-processing functions are applied to the oracle matrix of word embedding algorithms.

\section{Preliminaries} \label{qq}
This section presents a basic procedure for the dimension selection of word embedding.
A static word embedding algorithm is chosen to obtain an word embedding matrix $X \in \mathbb{R}^{n \times k}$, where $X$ is composed of $n$ vectors, and word $w_{i}$ is represented by vector $v_{i} \in \mathbb{R}^{k}$, $i \in [n]$. Generally, dimension selection aims to find a proper dimension $k$ with a good performance-efficiency trade-off.

There are two existing dimension selection methods: PIP-based and PCA-based. PIP can directly compute the dimension without training any word embedding, while PCA-based needs to train a word embedding before dimension selection, which leads to much more computational cost. So, the PCA-based method owns a worse performance-efficiency trade-off than the PIP-based method. In this paper, we work on dimension selection without training any word embedding. 

Following previous works \cite{levy2014neural}, we leverage oracle matrix for dimension selection, which can be obtained through matrix decomposition without any training process.
Formally, for a static word embedding algorithm with two hyper-parameters ($\alpha$ and dimension $k$) and a signal matrix $M$ (i.e., the co-occurrence matrix), an \textbf{oracle matrix} $X$ is derived from matrix $M$ with the form $X=f_{\alpha, k}(M) \leq U_{\cdot, 1: k} D_{1: k, 1: k}^{\alpha}$, where $M=U D V^{T}$ is its SVD obtained by matrix factorization \cite{levy2015improving}. 

To implement the dimension selection, we divide the training corpus into two identical partitions $C_{1}$ and $C_{2}$, and then obtain the co-occurrence matrices $M_{1}$ and $M_{2}$ on each partition, respectively. Correspondingly, we can have two oracle matrices $X$ and $\hat{X}$. Then, a common way to select a proper dimension is used: define a proper distance $d$, and then find $k$ that can minimize the distance $d(X,\hat{X})$. Notice that a distance ($\approx 0$) that guarantees the unitary-invariance of the two oracle matrices $X$ and $\hat{X}$ may serve as a proper distance for dimension selection \cite{yin2018dimensionality}, where unitary-invariance refers that two matrices are essentially identical if one can be transformed from the other by performing a unitary operation (e.g., a simple rotation), and unitary-invariance defines the equivalence class of the oracle matrices.


\section{Is word frequency influential in dimension selection?}\label{qqq}
Although previous works \cite{mu2018all,gong2018frage} have pointed out that word frequency may bring negative consequences to word embeddings, the impact of word frequency on the dimension selection has been overlooked by previous studies on dimension selection  \cite{yin2018dimensionality,wang2019single}. Therefore, we explore a question in this section: \textbf{does the word frequency influence the dimension selection for word embeddings?}.  

We first prepare an experiment of extreme word frequency on two tasks: context-unavailable (WordSim353 dataset, \citealp{finkelstein2001placing}) and context-available (SST-2 dataset, \citealp{socher2013reasoning}). In the experiment, the following two kinds of dimensions, $k^{*}$ and $k^{+}$, are used. Generally, word embedding with $k^{+}$ consistently outperforms the one with $k^{*}$, and the performance gap between $k^{+}$ and $k^{*}$ reflects the effectiveness of the dimension selection method.
\begin{itemize}
    \item Criterion-selected dimension $k^{*}$: Given a dimension selection method, $k^{*}$ with the minimal distance $d$ is selected as the criterion-selected dimension. 
    \item Empirically optimal dimension $k^{+}$ (Optimal): Given a word embedding algorithm, a word embedding is trained for each dimension in a grid search (i.e., the dimension is selected from 50 to 1000 with an increment step of 2). Then, each word embedding is evaluated on the benchmark, and the one achieving the highest performance is selected as the empirically optimal dimension $k^{+}$.
\end{itemize}

\begin{table}[!h]\label{freq}\small
\centering
\begin{tabular}{@{}c|c|cccc@{}}
\toprule
Dataset          & Criterion     & 0 & 3\%    & 5\%   & 10\%   \\ \midrule
\multirow{3}{*}{SST-2}    & $k^{+}$ (Optimal)              & 81.9 & 81.4 & 81.5 & 81.8 \\
                          & $k^{*}$ (PIP)                  & 80.0 & 79.5 & 79.2 & 78.7  \\
                          & $k^{*}$ (PCA)                  & 79.8 & 79.5 & 79.0 & 78.5  \\ \midrule
\multirow{3}{*}{WordSim353} & $k^{+}$ (Optimal)              & 69.1 & 69.2 & 69.1 & 69.1  \\
                          & $k^{*}$ (PIP)                  & 65.5 & 64.4 & 63.9 & 63.5  \\
                          & $k^{*}$ (PCA)                  & 64.4 & 64.0 & 63.5 & 63.2  \\ \bottomrule
\end{tabular}
\caption{Performance on the two tasks with extreme word frequency. The embedding are generated by Word2Vec based on a dimension selection criterion (PIP, PCA or grid search). `$10\%$' column means adding $10\% \cdot N$ duplicate sentences into the training corpus.}
\label{aaaa}
\end{table}

Given the training corpus, Gigaword 5 corpus which contains $N$ sentences, to obtain varying word frequencies, we randomly select a sentence (length is larger than 20, which will not affect the overall fluency of corpus considering the WordVec window size is 5) and duplicate it for $3\% \cdot N$, $5\% \cdot N$, and $10\% \cdot N$ times, and then merge them with the corpus. The results are listed in Table~\ref{aaaa}. As word frequency becomes extreme, the performance gap between $k^{*}$ and $k^{+}$ enlarges, indicating that extreme word frequency significantly hurts the effectiveness of dimension selection methods. Thus, it is necessary to diminish the influences of word frequency during dimension selection.

\section{Methodology}
\subsection{Overview}
The formulation of our proposed distance, mixed pairwise distance (MPD), is defined as follows:
\begin{equation}\label{eq1}
MPD(X, \hat{X})=\sqrt{d_{r}(X, \hat{X})\cdot d_{p}(X, \hat{X})}
\end{equation} 
where $X$ and $\hat{X}$ correspond to the two oracle matrices mentioned in Sec~\ref{qq}. Concretely, MPD consists of two distances: primitive relative distance $d_{r}$ and post relative distance $d_{p}$. Based on the distance criterion, an MPD-based dimension selection method is developed to select a proper dimension for static embedding algorithms.

\subsection{Primitive Relative Distance}
Definition~\ref{def1} gives the definition of the primitive relative distance $d_{r}$, and such a definition can guarantee the unitary-invariance of $X$ and $X^{T}$. Moreover, unless specifically stated, the matrix norms used in the paper are \textbf{Frobenius} norms.

\begin{definition}\label{def1}
The primitive relative distance $d_{r}$ between the oracle matrices $X \in \mathbb{R}^{n \times k}$ and $\hat{X} \in \mathbb{R}^{n \times k}$ is defined as follows:
\begin{equation}\label{eq2}
d_{r}\left(X, \hat{X}\right)= \frac{\left\|X X^{T}-\hat{X}\hat{X}^{T}\right\|^{2}}{\left\|X X^{T}\right\|\left\|\hat{X} \hat{X}^{T}\right\|}
\end{equation}
where the denominator $||X X^{T}|| ||\hat{X} \hat{X}^{T}||$ is a scaling term. 
\end{definition}

Then, the relationship between $d_{r}$ and unitary-invariance is presented as follows. According to Theorem~\ref{th2}, if $d_{r}$ is close to 0, the two matrices ($X$ and $\hat{X}$) are extremely unitary-invariant. In other words, when the numerator $||X X^{T}-\hat{X}\hat{X}^{T}||^{2}$ in Eq~\ref{eq2} is 0, $X$ and $\hat{X}$ can be regarded identical \cite{smith2017offline,artetxe2016learning}. Therefore, the definition of $d_{r}$ can guarantee that when $d_{r}$ is close to 0, the unitary-invariance of $X$ and $X^{T}$ exists. Thus, the optimal dimension which minimizes $d_{r}$ is a proper dimension for word embedding, as presented in Sec~\ref{qq}.




\begin{theorem}\label{th2}
Suppose $\left\|d_{r}(A,B)\right\| \approx 0$, then $A$ and $B$ are unitary-invariant (i.e., existing an unitary matrix $T$, $B \approx A T$).
\end{theorem}
The proof is deferred to Appendix~\ref{proof1}.

\subsection{Post Relative Distance}
In order to overcome the word frequency problem, the post-relative distance $d_{p}$ is defined, as shown in Definition~\ref{def3}, where $F(\cdot)$ is a post-processing function on the oracle matrices. Notice that the post-processing function in $d_{p}$ can be any post-processing method which has been used on word embedding.

\begin{definition}\label{def3}
Given a post-processing function $F$, the post-relative distance $d_{p}$ between the oracle matrices $X \in \mathbb{R}^{n \times k}$ and $\hat{X} \in \mathbb{R}^{n \times k}$ is defined as follows:
\begin{equation}
d_{p}\left(Y, \hat{Y}\right)= \frac{\left\|X Y^{T}-\hat{Y}\hat{Y}^{T}\right\|^{2}}{\left\|X Y^{T}\right\|\left\|\hat{Y}\hat{Y}^{T}\right\|}
\end{equation}
where $Y=F({X})$ and $\hat{Y}=F(\hat{X})$.
\end{definition}

Specifically, we transfer the success of the post-processing function on word embedding matrices to oracle matrices. Post-processing functions essentially `normalize' a word embedding through de-emphasizing the impact of some words. Since the oracle matrices, $X$ and $X^{T}$, are also influenced by the word frequency, we believe that the application of the post-processing function to the oracle matrices can also diminish such an influence.

Moreover, the post-relative distance is essentially the same as the primitive relative distance, except the usage of post-processing function $F$. Similar to the discussion of $d_{r}$, $d_{p}$ also possesses the unitary-invariance.

\subsection{Combination}
It is also worth mentioning that eliminating the impact of word frequency is not always beneficial for word embedding, because some crucial information in a word embedding is inevitably eliminated \cite{gong2018frage}. So, we combine $d_{r}$ and $d_{p}$ with a geometric mean in MPD.

\begin{table*}[!h]
\centering
\begin{tabular}{@{}c|c|cccccc@{}}
\toprule
Embedding           & Criterion & WS    & MTurk & MC    & MEN   & RG    & RW    \\ \midrule
BERT                      & None                & 41.15 & 32.89 & 49.13 & 41.23 & 34.66 & 13.15 \\ \midrule
\multirow{4}{*}{Glove}    & Optimal             & 69.12 & 58.78 & 70.20 & 73.75 & 76.95 & 33.55 \\
                          & PIP                 & 65.48 & 55.48 & 67.19 & 71.56 & 74.32 & 31.49 \\
                          & PCA                 & 64.44 & 55.23 & 66.82 & 71.09 & 73.56 & 31.67 \\
                          & MPD                 &\bf 66.89 &\bf 57.02 &\bf 68.66 &\bf 72.92 &\bf 75.23 &\bf 32.91 \\ \midrule
\multirow{4}{*}{Word2Vec} & Optimal             & 68.03 & 58.12 & 63.96 & 70.27 & 70.01 & 25.43 \\
                          & PIP                 & 64.68 & 55.63 & 62.19 & 68.45 & 68.76 & 23.12 \\
                          & PCA                 & 64.56 & 55.91 & 62.03 & 68.70 & 69.04 & 22.41 \\
                          & MPD                 &\bf 67.14 & \bf 57.11 & \bf62.88 & \bf69.79 &\bf 69.54 &\bf 24.64 \\ \bottomrule
\end{tabular}
\caption{The performance on word similarity using different word embeddings. Except the dynamic embedding produced by BERT, other word embeddings are generated by a static word embedding algorithm with a dimension selected by a criterion (PIP, PCA, MPD and a grid search); The number is 100 $\times$ $\rho$ (Spearman correlation); `\textbf{Bold}' numbers indicate the best performance except `Optimal'.}
\label{table1}
\end{table*}

\section{Experiment}
This section carries out three sets of experiments to comprehensively analyze our MPD-based dimension selection. The first two sets of experiments evaluate the effectiveness of our MPD-based dimension selection on context-unavailable and context-available tasks, respectively. The third set of experiments compares the efficiency-performance trade-off of different dimension selection methods. 

Specifically, two popular static word embedding algorithms (Word2Vec\footnote{We choose the CBOW version.} and GloVe) are chosen to train word embeddings and both of them use Gigaword 5 as the training corpus. For the MPD-based dimension selection method, two commonly-used post-processing methods (CN and ABTT) are used, and CN is the default one. Moreover, for each experiment, a \textbf{criterion-selected dimension} $k^{*}$ and an \textbf{empirically optimal dimension} $k^{+}$ are computed, whose definitions can be found in Sec~\ref{qqq}. 

\paragraph{Practical Settings}
Considering that SVD decomposition on large matrices needs extremely large memory resources, we reduce the vocab size in word embedding to 25,000, which can be handled by a 16GB-memory device. Moreover, previous works  \cite{yin2018dimensionality} have showed that as long as the vocab size is beyond a threshold (i.e., 10,000), the performance will not degrade much even when the vocab size is reduced.

\subsection{Context-unavailable Tasks}
\begin{table}[h]
\centering
\begin{tabular}{@{}c|c|c@{}}
\toprule
Dataset          & Embedding     & Human Score   \\ \midrule
\multirow{4}{*}{Chinese}    & BERT              & 0.79  \\ 
                          & $k^{*}$ (PIP)                  & 1.12   \\
                          & $k^{*}$ (PCA)                  & 1.08  \\
                          & $k^{*}$ (MPD)              & \bf1.38  \\
                          \midrule
\multirow{4}{*}{English} & BERT              & 0.68  \\ 
                          & $k^{*}$ (PIP)                  &1.12  \\
                          & $k^{*}$ (PCA)                  &1.05   \\
                          & $k^{*}$ (MPD)              & \bf1.29 \\
                          \bottomrule
\end{tabular}
\caption{Performance (Human score) on semantic expansion using different word embeddings, which are generated by GloVe with PIP, PCA or MPD. Particularly, empirically optimal dimension $k^{+}$ (Optimal) is not used because it is selected according to the final performance and such a selection is not plausible on semantic expansion since hundreds of times of human annotation are required.}
\label{tablese}
\end{table}

\paragraph{Experiment Settings}
We conduct experiments on the following two tasks in which contexts are unavailable. 
\begin{itemize}
    \item Word similarity: The quality of each embedding is evaluated on the following five benchmarks: WordSim353 \cite{finkelstein2001placing}, MTurk777 \cite{halawi2012large}, MC \cite{miller1991contextual}, MEN \cite{bruni2014multimodal}, RG \cite{rubenstein1965contextual}, and RW \cite{luong2013better}. Each dataset contains a list of word pairs with human scores. The performance is measured by the correlation between the cosine similarities of word vectors and human scores.
    \item Semantic expansion (SE): The Chinese and English entity list in TexSmart \cite{liu2021texsmart} is selected. Then, a word embedding trained with a selected dimension is applied to each entity to retrieve its related entities. Finally, the top five expansion results of each input entity are judged by human annotators in terms of quality and relatedness. Each result is annotated by two annotators, and a label of 2, 1 or 0 is assigned, corresponding to highly related, slightly related, and not related, respectively.
\end{itemize}

\paragraph{Results}
The performances of different dimension selection methods on word similarity and semantic expansion are listed in Table~\ref{table1} and Table~\ref{tablese}, respectively. First of all, we can see BERT \cite{devlin2018bert} performs catastrophically on context-unavailable tasks since the dynamic embedding generation is based on contextual learning. Then, MPD outperforms the baselines (PIP and PCA) in the word similarity tests because the post-processing function in MPD can diminish the impact of word frequency on the oracle matrices. Finally, MPD outperforms the baselines on semantic expansion because of the incorporation of the post-processing function, which enables the dimension selection method to neglect redundant components and capture word semantics better.



\subsection{Context-available Tasks}
\paragraph{Experiment Settings}
Since it is often a case that the success on the context-unavailable tasks cannot be well transferred to context-available tasks \cite{schnabel2015evaluation}, we conducted experiments that directly compare the effectiveness of different dimension selection methods on various downstream NLP tasks with contexts. The tasks can be divided into the following two kinds:

\textbf{Single-sentence tasks}: (1) Text classification: the movie review (MR) \cite{pang2005seeing}, SST-5 \cite{socher2013reasoning} and SST-2 datasets \cite{socher2013reasoning}; (2) Linguistic acceptability: CoLA  \cite{warstadt2019neural} dataset; (3) Dialogue state tracking: Wizard-of-Oz383 (WOZ) dataset \cite{wen2017network}.

\textbf{Sentence-pair tasks}: (1) Sentence paraphrase: MRPC \cite{dolan2005automatically} and QQP \cite{wang2018glue} dataset; (2) Semantic textual similarity: STS-B dataset \cite{cerasemeval}; (3) Language inference: MNLI dataset \cite{williams2018broad}.


Specifically, for the two single-sentence tasks (text classification and linguistic acceptability), a sentence is encoded with a static word embedding and the resulting embedding vector is fed to a Bi-LSTM classifier. For another single-sentence task, dialogue state tracking, a deep-neural-network-based model, Neural Belief Tracker (NBT) \cite{mrksic2018fully}, is chosen. For the three sentence-pair tasks, two sentences are encoded independently to produce two embedding vectors $u$ and $v$, and then vector $[u; v; |u - v|]$ is fed to a MLP-based classifier. Notice that the performances on these tasks can reflect whether a selected dimension is proper for static word embedding algorithms, but such performances still lag behind the ones using BERT. For the context-available tasks, it is not fair to compare PLMs (e.g., BERT) and static word embedding algorithms, because the former leverages contexts and the latter does not. Therefore, we only compare the four dimension selection methods (PIP-based, PCA-based, MPD-based, and grid search) in the experiments. 

\begin{table*}[h]\small
\centering
\begin{tabular}{@{}c|ccccc|ccccc@{}}
\toprule
Dataset                 & \multicolumn{5}{|c|}{Word2Vec}           & \multicolumn{5}{c}{GloVe}                                  \\ \midrule
          & PIP&PCA    & MPD(ABTT)  & MPD(CN)   & Optimal & PIP&PCA      & MPD(ABTT)& MPD(CN)           & Optimal \\ \midrule
MR        & 63.6 & 63.2  & {70.6}  & \textbf{70.9} & 71.4  & 65.0 & 69.4 & {71.3}  & \textbf{72.0} & 72.5  \\ \midrule
SST-2     & 80.0 & 79.8   & {81.6}  & \textbf{81.6} & 81.9 & 86.0 & 88.5 & \textbf{90.0} & {89.8} & 90.2  \\ \midrule
SST-5     & 32.4 & 32.0  & {38.0}  & \textbf{38.2} & 38.7  & 34.4 & 35.4 & \textbf{39.4} & {39.2} & 39.7  \\ \midrule
CoLA        & 4.3 & 4.4  & \textbf{5.6} & {5.5} & 5.6  & 14.4 & 14.3 & {15.6} & \textbf{15.6} & 15.8  \\ \midrule
WOZ       & 72.5 & 72.7  & \textbf{83.9} & {83.4} & 85.6  & 74.6 &79.4 & {86.5} & \textbf{90.2} & 92.4 \\ \midrule
MRPC & 79.4 & 79.3   & {80.6} & \textbf{80.9} & 80.9  & 80.0 &80.2 & {81.2} & \textbf{81.3} & 81.5  \\ \midrule
QQP   & 53.9 & 53.6  & {56.1}  & \textbf{56.2} & 56.6  & 60.4 &60.5 & {62.4}  & \textbf{62.7} & 62.9  \\ \midrule
STS-B    & 56.6 & 56.5  & {58.6}  & \textbf{58.8} & 59.0  & 58.0 &58.7  & {60.0}  & \textbf{60.2} & 60.4  \\ \midrule
MNLI & 54.4 & 54.0  & {57.0}  & \textbf{57.7} & 58.0  & 67.9 &67.6 & \textbf{71.5} & {71.4} & 71.6   \\ \bottomrule
\end{tabular}
\caption{The performance on the context-available tasks. Rows (e.g., `WOZ') represent the used datasets; Column `PIP', `PCA', `MPD' and `Optimal' represent PIP $k^{*}$, PCA $k^{*}$, MPD $k^{*}$ and empirically optimal dimensions $k^{+}$, respectively; Column `ABTT' and `CN' represent the used post-processing function in MPD, respectively.}
\label{table2}
\end{table*}


\paragraph{Results}
The experimental results of the six downstream NLP tasks are reported in Table~\ref{table2}. As shown in Table~\ref{table2}, the dimension  selection method using post-processing (i.e., Column `ABTT' and `CN') significantly outperform the ones without using post-processing (i.e., Column `PIP' and `PCA'). This shows the necessity of introducing post-processing to the dimension selection. Moreover, the MPD-based dimension selection achieves competitive performance with the optimal ones (`Optimal'). For example, on the 'WOZ' task, compared to PIP, the accuracy score based on MPD(ABTT) increases from $72.5\%$ to $83.9\%$, which is closer to the optimal performance ($85.6\%$). This means that MPD-based dimension selection works effectively for downstream tasks with contexts. 

Then, as illustrated in Table~\ref{table2}, the performance of MPD changes as the used post-processing method changes. Except the `WOZ' task, MPD(CN) usually performs on a par with MPD(ABTT). This indicates that the selection of post-processing functions does not affect much the MPD-based dimension selection. 


\subsection{Efficiency-performance Trade-off}
In this part, we compare the efficiency-performance trade-off among grid search and three dimension selection methods (PIP-based, PCA-based, and MPD-based) using Word2Vec, where efficiency is measured by time and performance is evaluated by the average per-task performance. Ideally, we hope a method could perform better and remain computationally efficient.
Specifically, the efficiency evaluation of grid search is dependent on grid granularity because a coarser grid search could save time at the cost of performance loss. Thus, in this experiment, we choose the dimension bound as $[50, 1000]$ and use three grid searches with three increments (i.e., 2, 5 and 10), respectively. The three grid searches are denoted as `GS-2',`GS-5', and `GS-10'. The efficiency-performance results are listed in Table~\ref{tableb}.  

In Table~\ref{tableb}, for the three grid searches, `GS-10' performs most closely to MPD but takes 14.01x time. Although `GS-2' and `GS-5' achieve slightly better performances than MPD, they take 56.04x and 28.02x computational cost, which is a disaster efficiency-performance trade-off. In the other hand, compared to PCA, MPD achieves $1.7\%$ performance improvement and costs only $17.1\%$ computation time of PCA. Compared to PIP, MPD has a similar computation cost, but achieves a $1.3\%$ performance increase. Therefore, the MPD-based method achieves the best efficiency-performance trade-off. 


\begin{table}[h]
\centering
\begin{tabular}{@{}c|cc@{}}
\toprule
Criterion & Performance & Computation time\\\midrule
GS-2 &59.5&56.04x\\
GS-5 &59.3&28.02x\\
GS-10 &59.0&14.01x\\
PIP &57.8&0.98x\\
PCA &57.3&10.85x\\
MPD &59.1&1.00x\\ \bottomrule
\end{tabular}
\caption{Comparison of efficiency-performance trade-off. \textbf{The whole MPD-based dimension selection procedure takes 6.3 min on one RTX2080 device.} Specifically, `5.85x' means 5.85 times of computational cost of MPD. }
\label{tableb}
\end{table}

\begin{table*}[!htbp]
\centering
\begin{tabular}{@{}c|cccc|cccc@{}}
\toprule
     &\multicolumn{4}{c|}{WS353}    &\multicolumn{4}{c}{MT771} \\ \midrule
    &MPD & Prim-D  & Post-D & Optimal &MPD & Prim-D  & Post-D & Optimal \\ \midrule
GloVe   &\textbf{0.66} &0.62 &0.65 &0.69 &\textbf{0.56}&0.55 &0.55 &0.58 \\ \midrule
Word2Vec  &\textbf{0.65} &0.60 &0.64 &0.68 &\textbf{0.55}&0.54 &0.55 &0.58 \\ \bottomrule
\end{tabular}
\caption{Ablation study on the context-unavailable tasks using GloVe and Word2Vec. `\textbf{Bold}' numbers indicate the best performance except `Optimal'.}
\label{table4}
\end{table*}

\begin{table*}[!h]
\centering
\begin{tabular}{@{}c|cccc|cccc@{}}
\toprule
Dataset         & \multicolumn{4}{c|}{ABTT}                    & \multicolumn{4}{c}{CN}                      \\ \midrule 
            & MPD           & Prim-D & Post-D & Optimal   & MPD           & Prim-D & Post-D & Optimal \\ \midrule 
MR              & \textbf{71.3} & 70.3 & 70.8 & 72.0    & \textbf{72.0} & 70.5 & 70.2 & 72.5    \\ \midrule 
SST-2         & \textbf{90.0} &89.2 &89.6 & 90.2                     & \textbf{89.8} & 89.1 & 88.9 & 90.2  \\ \midrule
SST-5         & \textbf{39.4} & 38.5 & 38.9 & 39.6    & \textbf{39.2} & 38.4 & 38.6 & 39.7    \\ \midrule 
CoLA         & \textbf{15.6} &15.0 &15.4 & 15.7         & \textbf{15.6} & 14.9 & 15.1 & 15.8  \\ \midrule
WOZ           &\textbf{86.1}&84.5&85.5&89.8  & \textbf{90.2} & 88.2 & 88.0 & 92.4 \\ \midrule
MRPC          & \textbf{81.2} & 80.4& 80.9 & 81.4          & \textbf{81.3}& 80.3 & 80.5 & 81.5  \\ \midrule
QQP      & \textbf{62.4} & 61.5 & 62.0 & 62.6         & \textbf{62.7} & 62.0 & 61.9 & 62.9  \\ \midrule
STS-B      & \textbf{60.0} &  59.2 &  59.8 & 60.2     & \textbf{60.2}& 59.6 & 59.7 & 60.4  \\ \midrule
MNLI      & \textbf{71.5} &  69.3 &  70.6 & 71.6 & \textbf{71.4} & 69.6 & 69.9 & 71.6      \\ \bottomrule
\end{tabular}
\caption{Ablation analysis on the context-available tasks using GloVe.}
\label{table5}
\end{table*}

\section{Ablation}
In the ablation studies, we principally answer two questions: (1) Is the distance combination in MPD effective? (2) Is the post-processing function effective in the case of extreme word frequency?

\subsection{Is the distance combination in MPD effective?}
To further explore the effectiveness of the combination of the two distances (the primitive relative distance $d_{r}$ and the post-relative distance $d_{p}$) in MPD, we add two extra dimension selection criteria: (1) Prim-D: $d_{r}$ (2) Post-D: $d_{p}$. Then, we compare the performances of the three dimension selection methods: MPD-based, Prim-D-based, and Post-D-based. 

Table~\ref{table4} and Table~\ref{table5} illustrate the performance of different selection methods on the context-unavailable and context-available tasks, respectively. As we can see, MPD obtains better results than the other two criteria, validating the effectiveness of the MPD which combines $d_{r}$ and $d_{p}$ through a geometric mean. The findings indicate that either completely preserving or diminishing the impact of word frequency is harmful to word embeddings, so a neutralized combination is proper. Although there may be other combinations of $d_{p}$ and $d_{r}$ with better performances, the distance combination method is not the focus of this paper.

\subsection{Is the post-processing function effective in the case of extreme word frequency?}
For the training corpus with extreme word frequency, the performances of different dimension selection methods are shown in Table~\ref{bbbb}. As we can see, the methods with post-processing functions (i.e., MPD and Post-D) perform better than or on a par with those without post-processing functions (i.e., PIP, MPD and Prim-D), demonstrating that the post-processing functions are effective in the case of extreme word frequency. 

\begin{table}[!h]\small
\centering
\begin{tabular}{@{}c|c|cccc@{}}
\toprule
Dataset          & Criterion     & 0 & 3\%    & 5\%   & 10\%   \\ \midrule
\multirow{6}{*}{SST-2}    & $k^{+}$ (Optimal)              & 81.9 & 81.4 & 81.5 & 81.8 \\
                          & $k^{*}$ (PIP)                  & 80.0 & 79.5 & 79.2 & 78.7  \\
                          & $k^{*}$ (PCA)                  & 79.8 & 79.5 & 79.0 & 78.5  \\
                          & $k^{*}$ (Prim-D)                  & 80.2 & 80.4 & 80.0 & 80.9\\
                          & $k^{*}$ (Post-D)                  & 80.6 & 80.9 & 79.9 & 80.5\\
                          & $k^{*}$ (MPD)                  &\bf 81.6 & \bf81.0 & \bf81.2 & \bf 81.3\\\midrule
\multirow{6}{*}{WordSim353} & $k^{+}$ (Optimal)              & 69.1 & 69.2 & 69.1 & 69.1  \\
                          & $k^{*}$ (PIP)                  & 65.5 & 64.4 & 63.9 & 63.5  \\
                          & $k^{*}$ (PCA)                  & 64.4 & 64.0 & 63.5 & 63.2  \\
                          & $k^{*}$ (Prim-D)                  & 62.3 & 62.2 & 61.2 & 61.0\\
                          & $k^{*}$ (Post-D)                  & 65.1 & 65.1 & 65.0 & 64.9\\
                          & $k^{*}$ (MPD)                  & \bf 66.9 & \bf66.6 & \bf66.5 & \bf66.5\\ \bottomrule
\end{tabular}
\caption{Ablation analysis on the two tasks with extreme word frequency. The experimental settings are the same as the Table~\ref{aaaa}. }
\label{bbbb}
\end{table}

\section{Conclusion}
In this paper, we investigate an overlooked problem, whether word frequency influences the automatic dimension selection for static word embedding algorithms, and discover that extreme word frequency can degrade existing dimension selection methods. According to the empirical finding, we propose MPD, a novel metric that combines primitive relative distance $d_{r}$ and the post relative distance $d_{p}$. Through $d_{p}$, post-processing functions are introduced into the dimension selection. Then, we develop an MPD-based dimension selection method that automatically selects a proper dimension for static embedding algorithms. Extensive experiments on context-unavailable and context-available tasks show that our MPD-based dimension selection achieves much better performance with a good efficiency-performance trade-off. 

\bibliography{anthology,custom}
\bibliographystyle{acl_natbib}

\appendix

\section{Proof of Theorem 1}\label{proof1}

\begin{proof}
Let $A=U D V^{T}$ and $B=X V Y^{T}$ be the SVDs. If we can prove $U \approx X$ and $D \approx V$, then $ B \approx A T$, and $T=V Y^{T}$ is the unitary matrix, which directly solves the proof. So we denote $D=\operatorname{diag}\left(d_{i}\right)$, $V=\operatorname{diag}\left(\lambda_{j}\right)$, and $d_{i}$ and $\lambda_{j}$ are the singular values of $A$ and $B$, respectively. The proof is shown as follows. 

Firstly, primitive relative distance $d_{r}$ defined in Eq. 2 is reformulated as Eq. 3:
\begin{equation}
\begin{aligned}
d_{r}\left(A, B\right)&= \frac{||A A^{T}-B B^{T}||}{C}\\ 
&=\frac{||U D^{2} U^{T}-X V^{2} X^{T}||}{C}
\end{aligned}
\end{equation} 
where $C=\frac{1}{||AA^{T}||||BB^{T}||}$. If $d_{r}(A,B) \approx 0$, then $||A A^{T}-B B^{T}|| \approx 0$. 

Secondly, let $x_{1}$ be the first column of $X$ (i.e., the singular vector corresponding to the largest singular value $\lambda_{1}$), and suppose $\lambda_{1} \geq d_{1}$. According to Eq. 10-12, $0 \leq \lambda_{1}^{2}-d_{1}^{2} \leq\left\|A A^{T}-B B^{T}\right\| \approx 0$. Then, $d_{1} \approx \lambda_{1}$ and $u_{1} \approx x_{1}$. 

\begin{equation}
\begin{aligned}
\left\|B B^{T} x_{1}\right\|-\left\|A A^{T} x_{1}\right\| & \leq\left\|\left(A A^{T}-B B^{T}\right) x_{1}\right\|_{op} \\
& \leq\left\|A A^{T}-B B^{T}\right\|
\end{aligned}
\end{equation} 
where $A A^{T}-B B^{T}$ is considered as an operator.

\begin{equation}
\left\|B B^{T} x_{1}\right\|=\left\|X V^{2} X^{T} x_{1}\right\|=\lambda_{1}^{2}
\end{equation}

\begin{equation}
\left\|A A^{T} x_{1}\right\|=||UD^{2}U^T x_{1}|| = d_{1}^{2}
\end{equation}

The proof can be applied to the other singular vectors of $X$ to get $U$ and $D$. Because $U \approx X$ and $D \approx V$, the unitary maxtrix $T$ is obtained, which completes the proof.
\end{proof}


\section{Limitation}
Since our MPD is specific to static word embedding, it can not be applied to pre-trained language models like BERT. Notice that BERT's `dimension' refers to the hidden state (768 in base). To dive into the relationship between dimension and BERT's performance, BERT is needed to be re-trained when the hidden state size changes. This is extremely difficult since re-training BERT needs massive computational resources, while re-training GloVe/Word2Vec can be easily handled. Besides, there are differences between various NLP tasks, but our dimension selection method lacks of task-specific designs, which will be explored in the future.

\end{document}